# How much more probable is "much more probable"? Verbal expressions for probability updates

Christopher Elsaesser and Max Henrion
Department of Engineering and Public Policy
Carnegie Mellon University
Pittsburgh, Pennsylvania 15213, USA

*Abstract*—Bayesian inference systems should be able to explain their reasoning to users, translating from numerical to natural language. Previous empirical work has investigated the correspondence between absolute probabilities and linguistic phrases. This study extends that work to the correspondence between changes in probabilities (updates) and relative probability phrases, such as "much more likely" or "a little less likely." Subjects selected such phrases to best describe numerical probability updates. We examined three hypotheses about the correspondence, and found the most descriptively accurate of these three to be that each such phrase corresponds to a fixed difference in probability (rather than fixed ratio of probabilities or of odds). The empirically derived phrase selection function uses eight phrases and achieved a 72% accuracy in correspondence with the subjects' actual usage.

## Introduction

A key characteristic for the acceptance of expert systems and other computer-based decision support systems is that they should be able to explain their reasoning in terms comprehensible to their users. Teach and Shortliffe (1984) found that physicians rated explanation an essential requirement for the acceptance of medical expert systems. Bayesian probabilistic inference has often been criticized as alien to human reasoning and so particularly hard to explain. However there have been a number of recent attempts to refute this accusation by the development of practical and effective systems for explaining Bayesian inference (Elsaesser, 1987, Norton, 1986, Speigelhalter, 1985). The work reported here is a part of such an attempt.

Bayesian reasoning usually uses numerical probabilities, but most people express preference for using natural language phrases, such as "probable", "very unlikely", "almost certain", and so forth. There has long been interest in developing empirical mappings between numbers and such probability phrases (e.g., Lichtenstein and Newman, 1967, Johnson, 1973, Beyth-Marom, 1982, Zimmer, 1983, Zimmer, 1985, Wallsten, et al., 1985). Most of these studies simply ask subjects to give the numerical probabilities they consider closest in meaning to selected phrases. This work has found considerable consistency in ranking of phrases between people, but moderate variability in the numbers assigned by different people. It has also found significant effect of the context on the numerical meaning assigned. Provided careful note is taken of the interperson and intercontext variabilities ("vagueness"), we may use such mappings from numbers to phrases to generate explanations automatically in probabilistic decision aids. The converse mapping from phrases to numbers may also be used as an aid to elicitation of expert uncertain opinions expressed as verbal probabilities. Sensitivity analysis of the effect of the vagueness should be used to check that this does not contribute unduly to vagueness in conclusions.

Hitherto all such work as been on absolute degrees of belief rather than changes in degrees of belief or probability updates. Since Bayesian inference is primarily about changes in probability, this seems an important lacuna. We have therefore chosen it as the focus of the study reported here. Specifically, our goal is to develop a mapping (a phrase selection function) that gives the relative probability phrase that best expresses a given change in probability. An example use of relative probability phrases are "little more likely" or "a great deal less likely". An example use might be as follows: Suppose the prior probability of Proposition A is 0.5 and evidence is presented to cause a revision to a posterior probability of 0.05. A Bayesian system might explain this thus:

"In light of the evidence, A is a great deal less likely."

## Hypotheses about Phrase Selection Functions

A *relative probability phrase selection function* gives a phrase for any change from a prior probability $p_1$ to a posterior probability $p_2$. It is a mapping from the unit square with dimensions $p_1$ and $p_2$ into a set of relative probability phrases.

f:[0,1] x [0,1] —> {relative probability phrases}

A phrase selection function effectively partitions the unit square into regions corresponding to specific relative probability phrases. (Figures 1, 2, and 3 show examples.) Our objective is to describe the phrase selection function for a fixed set of relative probability phrases that best fits our subjects' actual usage.

We should expect the partitions between regions to be monotonically increasing if the ordering of phrases is clear, but the actual shape of the partitioning curves is open to question. A priori, three alternative models seemed intuitively appealing:

$H_1$ *Constant probability ratio*: This phrase selection function, illustrated in Figure 1, is characterized by regions with partition lines of regions with $p_1$ proportional to $p_2$:

$$p_2 = c_i p_1$$

Note that $H_1$ exhibits range effects in probabilities. (A similar model was proposed by Oden (1977), but he was concerned with "relative belief" rather than relative probabilities.) One can draw an analogy between the constant ratio model and Fechner's law in psychophysics implying that a subjectively constant increment in the magnitude of a quantity is proportional to its its absolute magnitude.

$H_2$ *Constant probability difference*: This model is characterized by partition lines of the form:

$$p_2 = p_1 + c_i$$

Figure 2 allows a visual comparison with the other models. The constant probability difference model does *not* exhibit range effects in probability.

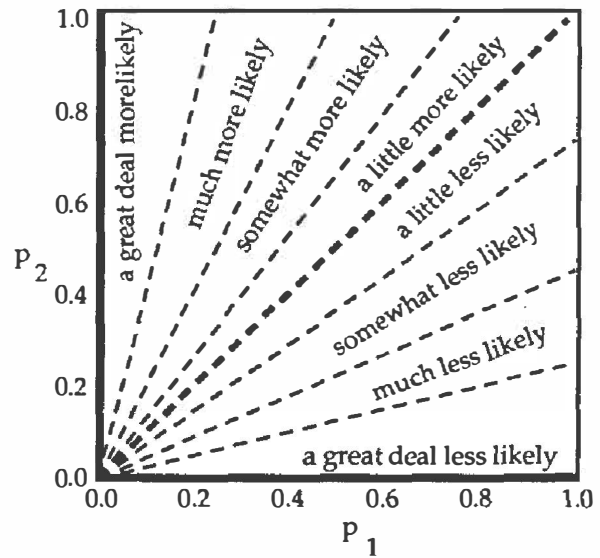

**Figure 1:** Hypothesis $H_1$ Constant probability ratio mapping from probability pairs to relative probability phrases

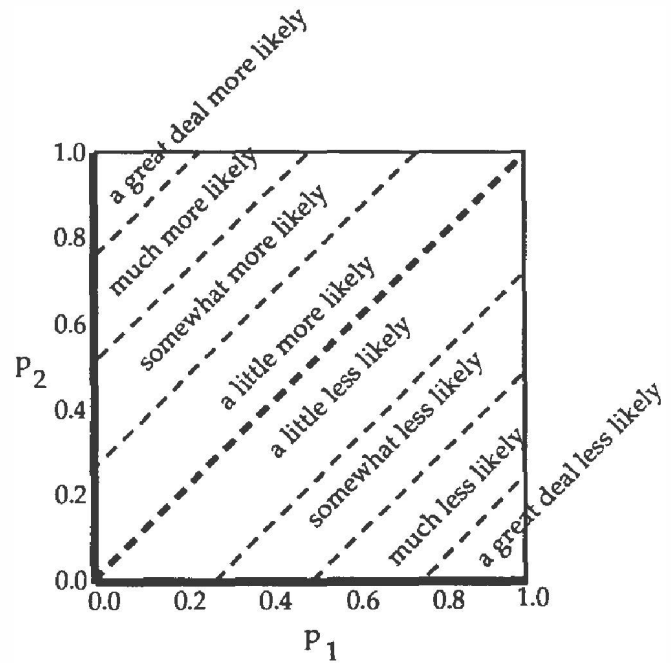

**Figure 2:** Hypothesis $H_2$: Constant probability difference

89

**$H_3$** *Constant odds ratio:* Partition can be represented as:

$$p_1(1-p_2) = c_i(1-p_1)p_2$$

$H_3$ is depicted graphically in Figure 3:

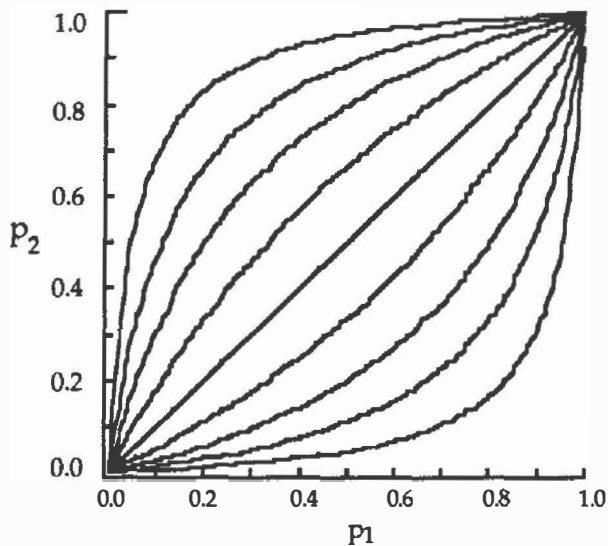

**Figure 3:** Hypothesis $H_3$ : Constant Odds ratio

$H_3$ exhibits range effects in probability but is linear in odds. $H_3$ resembles $H_1$ in a neighborhood of (0,0), resembles $H_2$ in the central region of the unit square, and exhibits range effects in the negation of the proposition of interest. All three models are symmetric about the line $p_1 = p_2$. But only $H_2$ and $H_3$ are symmetric for probabilities and their complements (i.e., about the diagonal $p_1 + p_2 = 1$.)

## Experimental Design

The experiment investigated how subjects assign a set of eight relative probability phrases to probability updates. We collected this empirical data to discover how to partition the unit square to best reflect the subjects' use of the phrases. This also enables us to compare the descriptive accuracy of these three hypotheses.

### Materials

Pairs of probabilities were selected randomly from the unit square, excluding extreme values, 0 and 100%, and pairs with equal values (no change). Thus, there were 98 * 98 -98 = 9506 possible pairs. Data was collected using questionnaires so that a large number of responses could be analyzed. Each subject was given 40 tasks, half with decreasing probabilities and half increasing. One practice question was given with the instructions. No subject received any duplicate tasks. The order of increasing and decreasing probabilities was randomized for each subject.

For each task, the subject was to mark the relative probability phrase he or she felt best described the given change from prior to posterior probability. Problems referred to abstract propositions to avoid context effects. Here is an example:

The probability of Event A had been estimated to be 5%, but new information caused the probability of A to be revised to 1%. Select the phrase which best completes the sentence:

In light of this new information, Event A is:
(_____) a great deal less likely.
(_____) much less likely.
(_____) somewhat less likely.
(_____) a little less likely.

The same set of alternative phrases were used for every question and presented in the same order to avoid inconsistency due to ordinal positioning effects (Hamm, 1988). Phrase sets were symmetric in their wording, with the words "more" and "less" substituted, depending on whether the prior and posterior probability used in the question increased or decreased.

### Subjects

Twenty-five members of the technical staff of the MITRE Corporation received questionnaires via inter-office mail, of which 19 (76%) gave useable responses. 58% of the respondents were male, 84% of respondents had some graduate education. All of the subjects were native English speakers.

### Response

747 probability pairs (prior, posterior) were interpreted by the 19 subjects. These pairs were collected into sets corresponding to the eight phrases. Phrase usage frequency ranged from 109 uses of "Quite a bit less" to 77 uses of "A little less," averaging 93 data points per phrase with a standard deviation of 10.7, indicating roughly uniform selection among the phrases.

## Analysis

Initial analysis employed regression to examine the relationship between $p_1$ and $p_2$ for each relative probability phrase i, using a polynomial (quadratic) relation:

$$p_2 = a_i + b_i\, p_1 + c_i\, p_1^2$$

For all phrases the linear terms were highly significant. Only for two phrases, "Quite a bit more" and "Quite a bit less" was the second-order term significant. This suggests that the constant odds ratio hypothesis $H_3$ is unlikely to be a good fit.

The primary goal was to find partition lines witch divide the unit square into eight regions corresponding to the eight relative probability phrases. Our objective was to maximize the agreement between the phrase selection function and the phrase choices by the subjects.

First, we plotted the probability pairs for each pair of adjacent phrases as scattergrams. Figure 4 shows a scattergram for two pairs of adjacent phrases. Next, we drew 20 equally-spaced diagonal lines perpendicular to the line $p_1 = p_2$. Between each pair of adjacent parallel diagonals we found a point that best partitioned the points corresponding to each phrase, i.e., minimizing the number of data points on the "wrong side" of the point. These partitioning points for each of the six pairs of adjacent phrases are plotted on Figure 5. Note that the phrases "a little more likely" and "a little less likely" are not considered adjacent, since they are separated by the phrase "equally likely" which we did not consider worth testing.

We then fit least-squares regression lines to each set of partitioning points. We found that quadratic and higher order terms yielded negligible improvement in fit (and more misclassification of phrases) over the linear term. The best fit straight lines are shown in Figure 5. Table 1 summarizes the partition lines resulting from this procedure, and their misclassification rates. Figure 6 shows the partition lines.

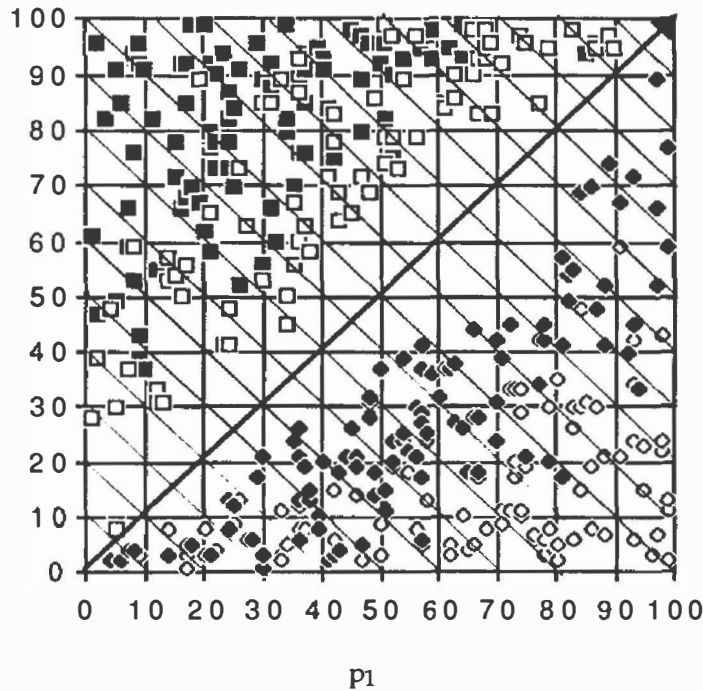

**Figure 4:** Scattergram for experiment data sets "Great deal more/less" and "Quite a bit more/less" with diagonal lines for determining partition lines.

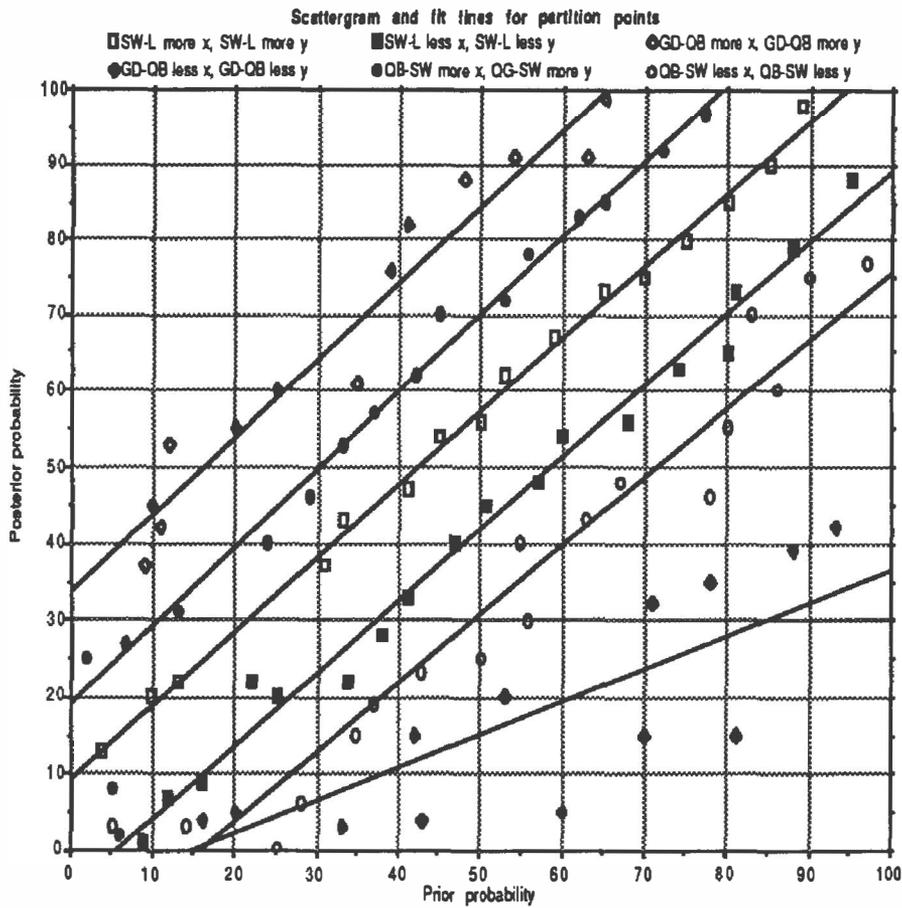

Figure 5: Least-squares fit lines for partition points.

Table 1: Summary of partition line analysis

| Phrase Pair | Partition line Slope | Intercept[1] | Misclassified points Above | Below |
|---|---|---|---|---|
| Great deal more | 1.01 | 32.0 | | 8 (8%) |
| Quite a bit more | | | 25 (30%) | |
| Quite a bit more | 1.05 | 18.0 | | 11 (14%) |
| Somewhat more | | | 17 (17%) | |
| Somewhat more | 0.98 | 9.4 | | 28 (28%) |
| Little more | | | 4 (4%) | |
| Little less | 0.96 | - 6.0 | | 9 (21%) |
| Somewhat less | | | 18 (18%) | |
| Somewhat less | 0.94 | -15.8 | | 28 (28%) |
| Quite a bit less | | | 24 (22%) | |
| Quite a bit less | 0.55 | -10.0 | | 24 (22%) |
| Great deal less | | | 22 (26%) | |

[1] Intercepts of the decreasing probability phrases on the prior axes are 6.3, 16.8, and 18.0.

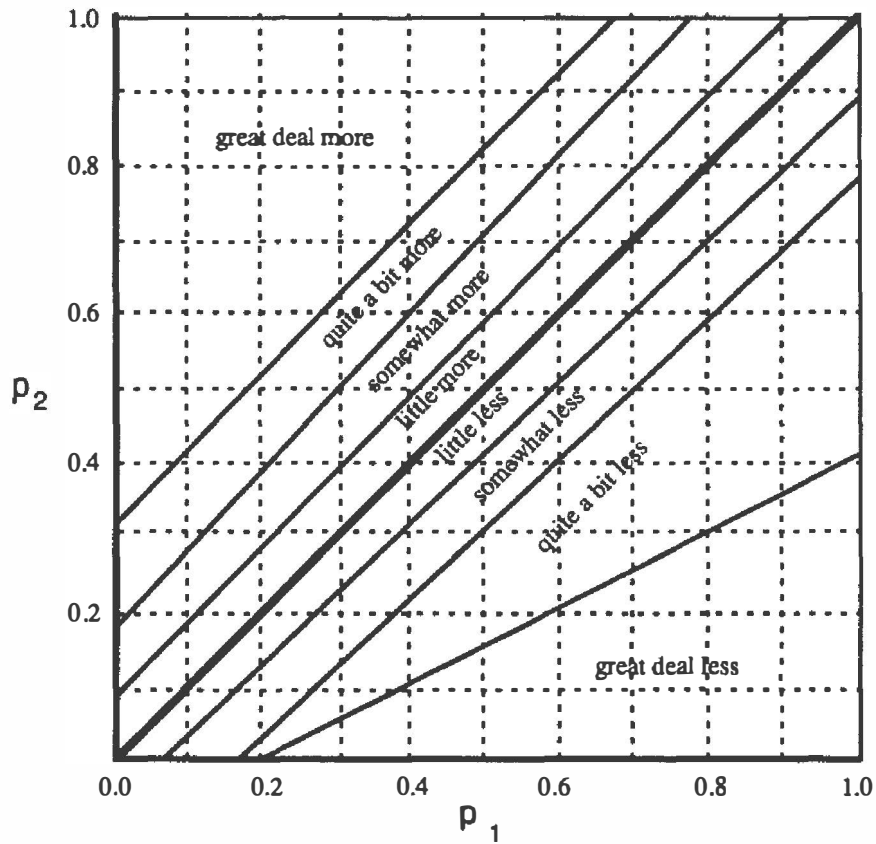

Figure 6: Phrase partition lines from non-parametric procedure

Overall these partitions match the subjects' choice of phrase in 72% of the cases. With one exception, namely the partition between "quite a bit less likely" and "a great deal less likely," these partitions are almost parallel. This provides support for model $H_2$, constant difference between $p_1$ and $p_2$, as a better description than $H_1$ and $H_3$. With the same exception, the partition lines are almost symmetric around the diagonal $p_1 = p_2$. This suggests a symmetry of treatment between prior and posterior probabilities.

## Conclusions

This paper examines one critical feature of a natural language explanation facility for Bayesian conditioning—the selection function for relative probability phrases. The investigation breaks new ground by extending the empirical study of probability numbers and phrases to probability updates and relative probability phrases. This preliminary investigation indicates that difference in probability is a better criterion for selecting a phrase to describe a change in probability, than ratio of probabilities or ratio of odds. The use of empirical partitions, for the eight phrases provided about 72% agreement with the choices of the subjects.

In this experiment we asked for selection of relative phrases given numerically specified probabilities. Further research might investigate the effect of specifying relative probability phrases and asking for posterior probabilities. Here we have examined only moderate probabilities (between 1% and 99%). The structure of the relation and appropriate phrases may look quite different for extreme probabilities (less than 1% or greater than 99%). Other questions for investigation include the selection of other relative probability phrases, and context effects.

Actually we doubt that such additional research will be likely to develop models that greatly improve on this performance using eight phrases for moderate probabilities due to the inherent variability among interpretations of such phrases. Nevertheless, we believe that this phrase selection function already provides sufficient resolution and agreement in usage to be useful as a basis for an explanation system for expert systems or other decision support systems using Bayesian inference. Of course, for any such scheme it will be important to bear in mind the vagueness due to the imprecision in usage and variability between people. No doubt context effects will provide additional variability as they have been shown to for absolute probability phrases. Any real system should be able to provide the actual numbers (or range of numbers) underlying any absolute or relative probability phrase used in the explanation for those users who require greater precision.

### Acknowledgements

Chris Elsaesser was supported in this work by the MITRE Corporation. Max Henrion was partly supported in this work by the National Science Foundation under grant number IRIS-8807061 to Carnegie-Mellon University.